\documentclass{article}

\usepackage{microtype}
\usepackage{graphicx}
\usepackage{subcaption}
\usepackage{booktabs} 

\usepackage{hyperref}
\usepackage{url}
\usepackage{xspace}
\usepackage{todonotes}
\usepackage{enumitem}
\usepackage{pifont}
\usepackage{wrapfig}
\usepackage{tcolorbox}
\usepackage{booktabs, tabularx, multirow}
\usepackage{framed}
\usepackage{listings}

\newcommand{\prompt}[1]{{\ttfamily #1}\xspace}
\newcommand{\sys}{TermiGen\xspace}
\newcommand{\agent}{BashAgent\xspace}

\newcommand*\user[1][1.5em]{$\vcenter{\hbox{\includegraphics[height=#1]{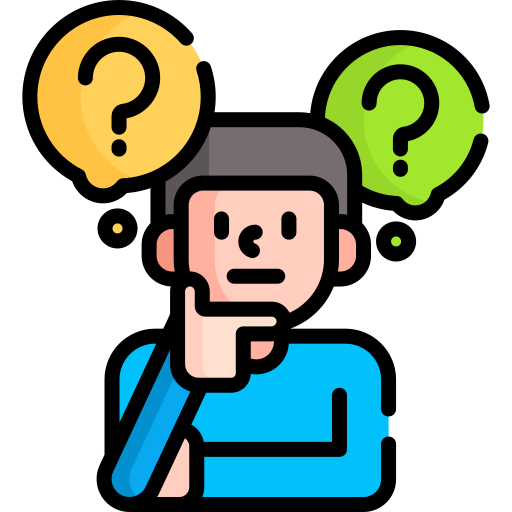}}}$}
\newcommand*\robot[1][1.5em]{$\vcenter{\hbox{\includegraphics[height=#1]{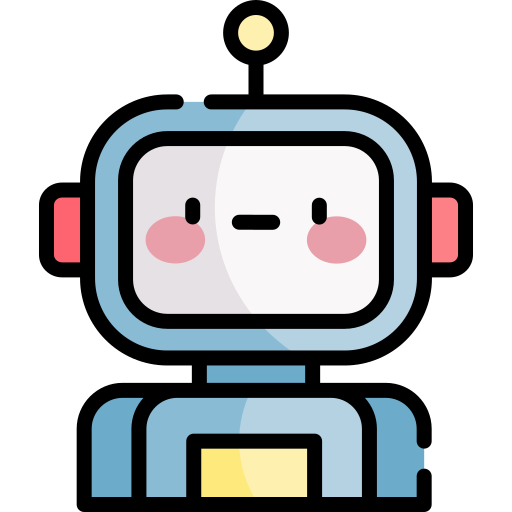}}}$}
\newcommand*\tool[1][1.5em]{$\vcenter{\hbox{\includegraphics[height=#1]{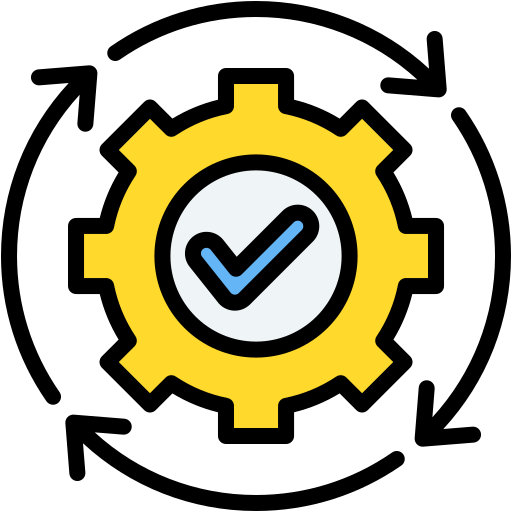}}}$}

\usepackage[preprint]{icml2026}

\usepackage{amsmath}
\usepackage{amssymb}
\usepackage{mathtools}
\usepackage{amsthm}

\usepackage[capitalize,noabbrev]{cleveref}

\theoremstyle{plain}

\theoremstyle{definition}

\theoremstyle{remark}

\icmltitlerunning{\sys: High-Fidelity Environment and Robust Trajectory Synthesis for Terminal Agents
}

\begin{document}


\twocolumn[
  \icmltitle{\sys: High-Fidelity Environment and Robust Trajectory Synthesis for Terminal Agents
}



  \icmlsetsymbol{equal}{*}

    \begin{icmlauthorlist}
    \icmlauthor{Kaijie Zhu}{ucsb}
    \icmlauthor{Yuzhou Nie}{ucsb}
    \icmlauthor{Yijiang Li}{ucsd}
    \icmlauthor{Yiming Huang}{ucsd}
    \icmlauthor{Jialian Wu}{amd}
    \icmlauthor{Jiang Liu}{amd}
    \icmlauthor{Ximeng Sun}{amd}
    \icmlauthor{Zhenfei Yin}{oxford}
    \icmlauthor{Lun Wang}{google}
    \icmlauthor{Zicheng Liu}{amd}
    \icmlauthor{Emad Barsoum}{amd}
    \icmlauthor{William Yang Wang}{ucsb}
    \icmlauthor{Wenbo Guo}{ucsb}
    \end{icmlauthorlist}
    
    \icmlaffiliation{ucsb}{UC, Santa Barbara}
    \icmlaffiliation{ucsd}{UC, San Diego}
     \icmlaffiliation{amd}{AMD}
    \icmlaffiliation{google}{Google}
    \icmlaffiliation{oxford}{University of Oxford}

    \icmlcorrespondingauthor{Kaijie Zhu}{kaijiezhu@ucsb.edu}

  \icmlkeywords{Machine Learning, ICML}

  \vskip 0.3in
]



\printAffiliationsAndNotice{}  

\begin{abstract}
Executing complex terminal tasks remains a significant challenge for open-weight LLMs, constrained by two fundamental limitations. 
First, high-fidelity, executable training environments are scarce: environments synthesized from real-world repositories are not diverse and scalable, while trajectories synthesized by LLMs suffer from hallucinations.
Second, standard instruction tuning uses expert trajectories that rarely exhibit simple mistakes common to smaller models. 
This creates a distributional mismatch, leaving student models ill-equipped to recover from their own runtime failures. 
To bridge these gaps, we introduce \textbf{\sys}, an end-to-end pipeline for synthesizing verifiable environments and resilient expert trajectories. 
\sys first generates functionally valid tasks and Docker containers via an iterative multi-agent refinement loop. 
Subsequently, we employ a \textit{Generator-Critic} protocol that actively injects errors during trajectory collection, synthesizing data rich in \textit{error-correction} cycles. 
Fine-tuned on this \sys-generated dataset, our \sys-Qwen2.5-Coder-32B achieves a $31.3$\% pass rate on TerminalBench.
This establishes a new open-weights state-of-the-art, outperforming existing baselines and notably surpassing capable proprietary models such as o4-mini.
Dataset is avaiable at \url{https://github.com/ucsb-mlsec/terminal-bench-env}.
\end{abstract}

\begin{figure}[t!]
    \centering
    \includegraphics[width=0.5\textwidth]{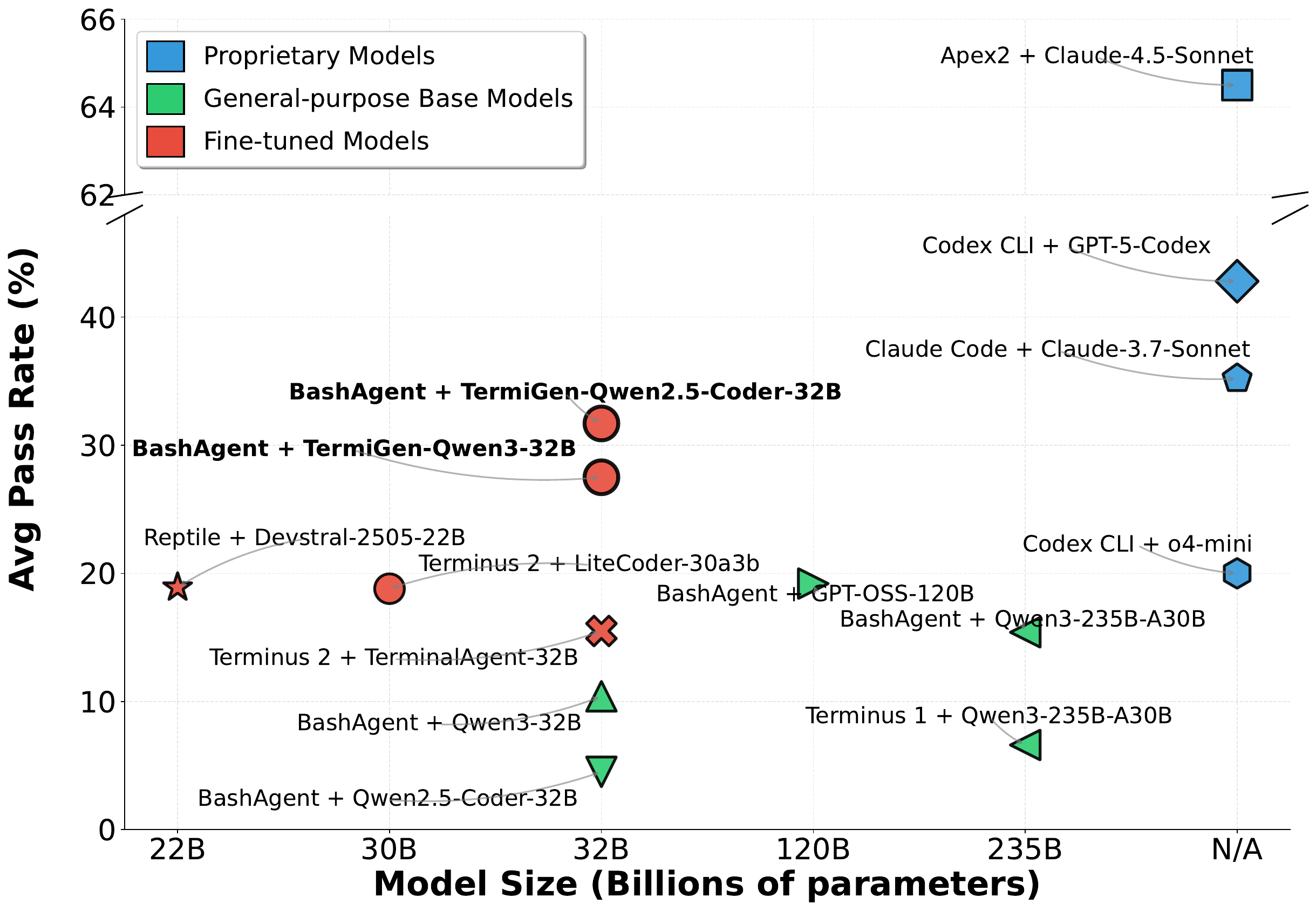}
    \caption{Average pass rate on Terminal Bench (\%) vs. model size. \sys models (bold) outperform other open-source baselines and approach proprietary system performance at 32B scale. 
    Blue refers to proprietary models, green refers to general-purpose base models, and red refers to fine-tuned models.  
    }
    \label{fig:tool_distribution}
    \vspace{-.1in}
\end{figure}

\section{Introduction}
\label{sec:intro}
Recent advancements have empowered LLMs with agentic scaffolding and tool-use capabilities, enabling them to tackle complex tasks through interaction with real-world environments~\citep{zhou2023webarena, jimenez2023swe, xie2024osworld, tbench2025}.
Among these domains, executing software engineering and system administration tasks via the terminal (i.e., terminal tasks) stands out as one of the most fundamental yet challenging frontiers~\cite{tbench2025}.
Distinct from standard code generation~\citep{chen2021evaluating, huang2024competition} or math 
reasoning~\citep{cobbe2021training}, terminal tasks pose two core challenges and have thus far been dominated by large proprietary models backed by extensive private training data.

\textbf{First, scalable construction of executable training environments remains a major bottleneck.}
Unlike individual code snippets, terminal tasks require a complete, executable specification of system architecture, file dependencies, and dynamic runtime status. 
Constructing such complex and verifiable environments necessitates substantial effort from domain experts. 
TerminalBench~\citep{tbench2025}, for instance, required extensive manual curation to produce merely 200 tasks.
Existing data synthesis approaches generally fall into two paradigms, each with significant limitations.
The first paradigm adapts existing software repositories (e.g., mining GitHub issues~\citep{yang2025swe}).
While grounded in real code, these sources often lack diversity and are biased towards debugging and issue resolving~\citep{liu2024large}.
They also require extensive manual effort to define verifiable success criteria for each task.
The second paradigm involves purely LLM-based synthesis, where models are prompted to generate hypothetical execution logs without access to a live runtime~\citep{chen2025scaling, team2025kimi}.
However, as our experiments reveal~(\cref{subsec:ablation}), this approach suffers from severe hallucinations: without grounded execution, the synthesizer invents plausible but technically incorrect outcomes (e.g., reporting successful installation despite dependency conflicts), which can mislead student models during training.

\textbf{Second, long-horizon terminal execution worsens exposure bias.}
Completing a terminal task often requires executing long command sequences where a single intermediate mistake can cascade into an irreversible failure. 
Standard distillation methods typically perform rejection sampling~\citep{mitra2024agentinstruct} on expert models to harvest optimal execution paths.
However, for long-horizon terminal tasks, we find that success crucially depends on the ability to detect, critique, and correct one’s own errors.
Since expert models rarely make simple mistakes, standard datasets suffer from a \textit{distributional mismatch}, leaving student models ill-equipped to handle runtime failures.
Consequently, student models suffer from a distribution shift during deployment, they frequently encounter trivial errors (e.g., dependency conflicts) that were absent during training, lacking the requisite supervision to recover from these failure states.

We aim to close the significant gap between open-weight agents and large proprietary models in the terminal tasks.
We approach this challenge from a data-centric perspective, i.e., an automated, scalable pipeline for high-fidelity environment synthesis is the foundational prerequisite for effective training.
Guided by this insight, we propose \textbf{\sys}, an end-to-end data generation recipe specifically designed for developing robust terminal agents.
At a high level, \sys bridges the data scarcity gap by first synthesizing verifiable tasks via a multi-stage framework, and subsequently employing a controlled error-correction mechanism to collect realistic expert trajectories that facilitate resilient learning.
Specifically, for environment synthesis, we employ a multi-agent system to decompose the generation process, ensuring coverage across diverse task categories.
Crucially, to eliminate the hallucinations common in simulation-based approaches, we introduce an interactive execution loop that validates environment correctness and task solvability within a real Docker container before data collection begins.
Following environment generation, we propose a novel trajectory collection protocol designed to mitigate the distributional mismatch in the standard distillation process.
To incentivize resiliency, we employ a \textit{Generator-Critic} architecture to generate error-correction trajectories in a controlled manner.
At each action step, we stochastically inject realistic faults into the trajectory and condition the agent to diagnose the failure and recover the system state subsequently.
This process yields training data rich in explicit \textit{error $\to$ diagnosis $\to$ correction} cycles, teaching the model how to recover from runtime mistakes.

We conduct extensive evaluations to validate our framework.
First, our \sys-Qwen2.5-Coder-32B model achieves a pass rate of \textbf{31.3\%} on TerminalBench, \textit{establishing a new state-of-the-art for open-weights models} and outperforming existing fine-tuning baselines like Reptile by over \textbf{12\%}.
Notably, it even surpasses capable proprietary models such as o4-mini with Codex CLI, demonstrating that high-quality data can enable smaller models to rival larger counterparts in specialized domains.
Second, we demonstrate that training on grounded, verifiable environments significantly outperforms simulation-based baselines.
Third, our ablation studies reveal that the inclusion of error-correction trajectories improves pass rates by a substantial margin compared to training solely on standard expert trajectories.
Last, we show that including error trajectories which fail to reach the task goal also improves performance.
To the best of our knowledge, \sys is the first end-to-end recipe for training terminal agents that integrates a high-fidelity, robust, and scalable data synthesis pipeline.

\section{Existing Works and Limitations}
\label{sec:rw}

\textbf{Environment for Agentic Training.}
At a high level, existing works for constructing agent environments fall into two primary categories. 
The first line of methods focuses on automatically constructing executable environments from existing code repositories, utilizing either a semi-automated pipeline~\citep{yang2025swe} or a fully automated pipeline~\citep{zhang2025swe}. 
However, the dependency on pre-existing projects constrains their generalizability to more diverse and unseen tasks and environment settings.
Another line of work employs LLMs as simulated environments to directly generate feedback~\citep{chen2025scaling, team2025kimi, li2025simia, wang2025llmsimulator}, thereby facilitating large-scale data synthesis and reinforcement learning. While simulation reduces reliance on real environments, the generated feedback can be inconsistent with actual execution (e.g., state-tracking errors or hallucinated effects), which may mislead agent training. 
In contrast, our framework generates task-specific Docker-based environments that support actual execution, ensuring that agents are trained in diverse, grounded, and realistic environments.

\begin{figure*}[t!]
    \centering
    \includegraphics[width=\textwidth]{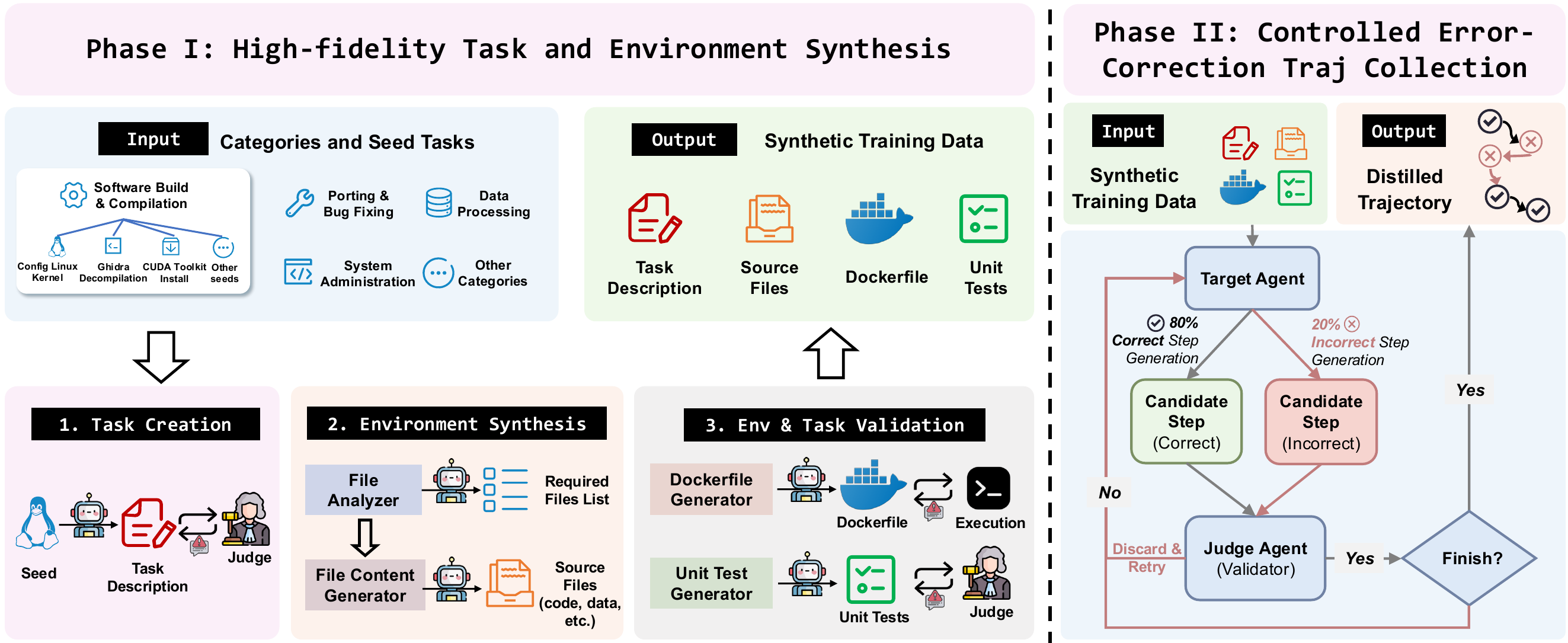}
    \caption{The overall pipeline of \sys. Phase I generates diverse, functionally valid tasks within Docker containers via iterative refinement~(\cref{subsec:env_synthesis}). Phase II synthesizes robust expert trajectories by actively injecting errors into the execution flow, enabling the model to learn error diagnosis and recovery~(\cref{subsec:trajectory_collection}).}
    \label{fig:overall}
    \vspace{-.1in}
\end{figure*}

\textbf{Agentic Supervised Fine-Tuning and Learning from Failure.}
Standard approaches for training agentic small models typically rely on SFT over expert trajectories distilled from frontier models~\citep{chen2023fireact, zeng2024agenttuning}.
While effective for cloning optimal behaviors, this paradigm suffers from \textit{exposure bias}~\citep{ross2011reduction, bengio2015scheduled}: since experts rarely make simple mistakes, the resulting datasets lack failure states and corresponding recovery phases.
Consequently, student models trained solely on expert trajectories become fragile, failing catastrophically when they inevitably encounter runtime exceptions.
To address this, recent works utilize execution feedback or leverage iterative self-refinement to guide self-correction~\citep{shinn2023reflexion, madaan2023self}.
However, these methods predominantly focus on \textit{inference-time} refinement rather than model training.
Although some efforts explicitly train on error-correction data~\citep{zheng2024opencodeinterpreter}, they are often limited to single-turn, static code repair tasks.
Our framework advances this direction by integrating active error injection into the trajectory synthesis pipeline, enabling the model to internalize the full \textit{error $\rightarrow$ diagnosis $\rightarrow$ recovery} cycle within complex, multi-turn terminal interactions.

\section{Key Technique}
\label{sec:method}

\subsection{Overview}
\label{subsec:overview}

\textbf{Technical Challenges.}
Training agentic small models for terminal agents has two unique challenges compared to standard code and math reasoning tasks.

\ding{182} \underline{Scarcity of Diverse, Executable Environments.} 
Unlike self-contained coding tasks (e.g., LeetCode), terminal tasks are \textit{environment-dependent}, requiring specific OS configurations and runtime dependencies. 
Constructing such environments manually is prohibitively expensive, forcing existing benchmarks to focus narrowly on code debugging while leaving critical domains like system administration, security forensics, and infrastructure management unaddressed.
Models trained on these limited tasks exhibit restricted reasoning capabilities for broader tasks and fail to utilize a diverse range of terminal operations.
Attempts to bypass this via \textit{simulated execution} face a fundamental \textit{fidelity gap}, as the synthesized trajectories may not represent the real execution traces and may contain hallucinations.
A typical issue is \textit{state inconsistency}.
For example, if an agent modifies a file (e.g., changing a threshold in \texttt{config.py}), simulators often fail to track this persistent change, hallucinating outcomes based on the original version. 
This necessitates a pipeline that ensures both \textit{executability} (to avoid hallucinations) and \textit{scalability} (to cover diverse domains).

\ding{183} \underline{Necessity of Error-correction Training.} 
Software and system-related terminal tasks typically require long action sequences (often $>20$ steps), where \textit{any single step can fail}.
However, standard expert demonstrations used for training are typically sparse in errors, as strong teacher models rarely make the trivial mistakes that smaller student models encounter. 
Consequently, agents trained on such data suffer from \textit{exposure bias}: they lack the resilience to correct runtime exceptions. 
When an error occurs during deployment, the resulting failure state (e.g., confusing stderr, corrupted file systems) represents an out-of-distribution (OOD) scenario relative to the training data.
Self-correction is particularly critical for such agentic tasks, as external tool calls induce state changes that are inherently more difficult to reverse than internal reasoning errors.
This necessitates training on \textit{imperfect expert demonstrations} that explicitly include error-correction sequences, which show not only correct execution paths, but also how to recognize mistakes and patch them through self-correction.

\textbf{Insight of \sys.}
In this work, we propose \sys, the first end-to-end framework for training effective terminal agents with the self-refinement capability. 
Our method consists of the following two phases, designed to address the aforementioned challenges.
\cref{fig:overall} shows an overview of \sys pipeline.

\underline{Phase I: High-fidelity Task and Environment Synthesis.} 
We implement a multi-agent framework to decompose the generation process into controllable stages.
Guided by a structured taxonomy (spanning $11$ categories across three tiers), we first instruct an LLM to generate distinct task seeds to ensure broad coverage across domains. 
These seeds are concise, high-level objective descriptions.
Subsequently, we employ a \textit{Proposer-Evaluator} architecture to expand these seeds into fully specified tasks (e.g., adding detailed input/output requirements). 
In this loop, the Evaluator validates the generated content against predefined criteria, providing feedback to guide iterative refinement.
We then leverage a structured workflow to translate these textual descriptions into executable Dockerfiles.
Given the complexity of these domains, ensuring environment correctness and task validity is non-trivial. 
We ensure data quality through a two-stage automated validation pipeline.
First, to guarantee environment executability, we implement an iterative Docker build loop where build failures are automatically diagnosed and corrected.
Second, to verify task solvability, we employ a \textit{Generator-Judge} framework: a Generator Agent synthesizes initial unit tests, which are then iteratively validated and refined by a Judge Agent to ensure they accurately capture task objectives and provide reliable execution feedback.

\underline{Phase II: Controlled Error-Correction Trajectory Collection.}
Following environment verification, we proceed to trajectory synthesis. 
To address the exposure bias inherent in standard distillation (as discussed in Challenge \ding{183}), we introduce an active error injection mechanism.
We employ a \textit{Generator-Critic} architecture that stochastically injects controlled failures sampled from a taxonomy of five common failure modes~(\cref{subsec:trajectory_collection}).
The generator operates under a mixed policy: at each step, it probabilistically determines whether to pursue the optimal path or intentionally deviate to commit a specific error (e.g., invoking non-existent arguments in a command).
A \textit{Critic Agent} validates the alignment between actions and intents, ensuring that injected errors trigger informative feedback (stderr) while correct steps meaningfully advance the task.
Crucially, whenever the policy dictates a return to generate a correct step, the generator is conditioned to diagnose the preceding failure and synthesize a corrective solution.
This process yields trajectories enriched with explicit \textit{error $\to$ diagnosis $\to$ recovery} cycles.

\subsection{High-Fidelity Task and Environment Synthesis}
\label{subsec:env_synthesis}

We now zoom into the technical details of our task and environment synthesis pipeline, which generates diverse, executable terminal tasks and environments at scale. 
The pipeline consists of three stages: 
1) \textit{Task Definition and Generation}, generating diverse and verifiable tasks; 
2) \textit{Automated Environment Synthesis}, generating required files and Docker containers; and 
3) \textit{Environment and Task validation}, where we iteratively test and validate the correctness of the environments and the resolvability of the tasks.
The workflow is shown in~\cref{fig:overall} Phase~I.

\textbf{Stage 1: Task Generation.}
To ensure comprehensive coverage, we construct a hierarchical taxonomy (Appendix~\cref{tab:taxonomy}) that spans the full spectrum of terminal usage: from low-level system operations to high-level application workflows.
The taxonomy is structured into three tiers comprising $11$ distinct categories:
1)~\underline{Tier 1: Infrastructure and core systems} focuses on low-level system interactions, the corresponding categories range from \textit{build systems} (e.g., resolving \texttt{gcc} dependency for Linux kernels) to \textit{system administration} (e.g., configuring Docker).
2)~\underline{Data and algorithm applications} covers data-centric workflows (e.g., merging parquet files), MLOps (e.g., resolving CUDA out-of-memory), and algorithmic reasoning tasks (e.g., implementing A* algorithm).
3)~\underline{Specialized development} targets advanced engineering scenarios such as software development (e.g., debugging PostgreSQL deadlock), scientific computing (e.g., fixing RDKit molecule sanitization), etc.
Guided by this taxonomy, we prompt an LLM to generate distinct task seeds, defined as concise, high-level abstracts such as ``\prompt{Resolving gcc -l library not found errors}'', aiming to ensure broad and non-overlapping coverage within each category.

Since these seeds are merely high-level directives, we employ a \textit{Task Proposal Agent} to instantiate each seed into a concrete definition.
The agent is prompted to synthesize a structured specification, comprising the detailed objective, input/output requirements, and success criteria.
However, unconstrained generation often yields tasks that are infeasible or non-reproducible (e.g., ``\prompt{optimize server latency to $<10$ms}'', which is hardware-dependent).
To mitigate this, we introduce a \textit{Task Evaluator Agent}, establishing a Proposer-Evaluator architecture for iterative refinement.
The Evaluator critiques task proposals based on three feasibility metrics (scored $1$--$5$):
1) \underline{Environment Complexity}: Prioritizing tasks using standard packages (e.g., \texttt{apt}, \texttt{pip}) over those requiring special hardware or obscure dependencies.
2) \underline{Data Generatability}: Ensuring input artifacts (e.g., logs, configs) can be synthesized by LLMs without requiring external production dumps.
3) \underline{Verification Determinism}: Favoring tasks verifiable via deterministic rules (e.g., \texttt{pytest}) rather than ambiguous state checks.
We enforce a strict quality threshold: only tasks scoring $>4$ in all dimensions are accepted.
Low-scoring proposals trigger a feedback-guided refinement loop (max $3$ rounds) until the criteria are met.

\textbf{Stage 2: Environment Synthesis.}
Given the detailed task description from Stage 1, we employ a sequential workflow where specialized agents collaborate to materialize the environment:
1)~\underline{Structural Planning:} 
A \textit{File Planner Agent} analyzes the description to decompose it into a concrete file system blueprint, outlining the directory structure and specific requirements for each artifact.
2)~\underline{Content Generation:} 
Guided by this blueprint, a \textit{Construct Agent} instantiates the actual content for each file separately (e.g., synthesizing faulty Python scripts, configuration logs, or Makefiles).
3)~\underline{Dockerfile Generation:} 
Finally, an \textit{Env Agent} generates a \texttt{Dockerfile} to encapsulate these artifacts. 
Crucially, the agent is instructed to explicitly resolve system-level dependencies (e.g., installing \texttt{libssl-dev} for compilation), ensuring the environment is fully self-contained.

\textbf{Stage 3: Environment and Task Validation.}
We note that the generated Dockerfiles often contain subtle errors (e.g., deprecated packages, version mismatches). 
We implement an iterative refinement loop: every generated environment is submitted to the Docker daemon. 
If Docker build fails, the \textit{stderr} log (e.g., \prompt{E: Unable to locate package}) is captured and fed back to \textit{Env Agent} as a prompt for correction. 
This cycle repeats (up to $N=5$ iterations) until the build succeeds. 
This process filters out non-executable configurations, ensuring that $100$\% of our training environments are functionally valid.

We further proposed a \textit{Unit Test Generation and Validation} process for verifying that the generated tasks are solvable.
We first instruct a \textit{Unit Test Generator Agent} to generate unit tests for each task.
However, generating reliable unit tests is inherently challenging, even more so than solving the tasks themselves~\citep{mundler2024swt, tang2026devops}.
A valid unit test must correctly distinguish ground-truth solutions (which pass) from plausible but incorrect attempts (which fail), requiring a precise understanding of success criteria and edge cases. 
This difficulty is compounded in our setting as we lack ground-truth solutions to verify against. 
Nevertheless, unit tests are critical for both trajectory collection (providing execution feedback) and reinforcement learning (defining reward signals). 
To address this, we further employ a \textit{Judge Agent} that validates each generated unit test through iterative refinement: it verifies that unit tests correctly capture task objectives, provide adequate coverage of success conditions, and execute without errors. 
Tests failing validation receive detailed feedback and are regenerated by the \textit{Unit Test Generator} until the \textit{Judge Agent} approves.

\subsection{Error-Correction Trajectory Collection}
\label{subsec:trajectory_collection}

Following the verification of environments, we proceed to trajectory synthesis. 
To mitigate the exposure bias discussed in Challenge~\ding{183}, we move beyond standard distillation by actively injecting failures into the training data.
We implement this via a \textit{Generator-Critic} framework (as shown in \cref{fig:overall} Phase II).
At each time step $t$, framework operates through three sequential steps.

\textbf{Step 1: Intent Sampling.}
To control the trajectory generation, we explicitly sample an \textit{intent signal} $I_t$ from a Bernoulli distribution parameterized by an injection rate $\epsilon$:
\begin{equation}
    I_t \sim 
    \begin{cases} 
    \mathcal{I}_{\text{correct}} & \text{with probability } 1 - \epsilon \\
    \mathcal{I}_{\text{error}} & \text{with probability } \epsilon
    \end{cases}
\end{equation}
where we set $\epsilon = 0.2$. 
If $I_t = \mathcal{I}_{\text{correct}}$, the agent aims to advance the task state $s_t$ efficiently.
We note that the term ``correct'' in this context does not necessarily imply that the step is entirely accurate. 
Rather, it signifies that the agent is attempting to advance the task, as opposed to deliberately introducing errors.
Conversely, if $I_t = \mathcal{I}_{\text{error}}$, the agent is instructed to commit a \textit{sophisticated error}.

\textbf{Step 2: Context-Aware Generation.}
Guided by the sampled intent, the \textit{Step Generator} produces a ReAct-style action $a_t$ conditioned on the current system state $s_t$ and task context.
Crucially, when $I_t = \mathcal{I}_{\text{error}}$, the agent is prompted to synthesize \textit{plausible} domain-specific mistakes rather than random noise.
To ensure comprehensive coverage, we taxonomize agent failures into five categories: 
\textit{1) Analysis Errors:} misinterpretation of environment states or data structures; 
\textit{2) Command Errors:} syntactic or formatting failures in tool execution; 
\textit{3) Hallucinations}, where agents assume the existence of absent tools or services; 
\textit{4) Requirement Violations:} neglecting explicit task constraints; 
\textit{5) Verification Failures:} lack of self-check steps before concluding tasks.
For example, in a system administration context, it might attempt a privileged operation without \texttt{sudo} (a common \textit{Permission Oversight}), ensuring the error reflects realistic user behavior.

\textbf{Step 3: Critic Validation.}
Finally, a \textit{Critic Agent} validates the alignment between the generated action $a_t$ and the intent $I_t$ to ensure data quality.
For \textit{error steps}, it filters out low-quality error injections by verifying that the action is syntactically plausible and will trigger \textit{informative feedback} (e.g., a specific stderr message) upon execution.
For \textit{optimal steps}, it confirms that the action effectively advances the task state.
Any step failing this validation triggers a regeneration loop guided by the \textit{Critic Agent}'s feedback.

\textbf{Handling Cumulative Errors and Recovery.}
The stochastic nature of our intent sampling enables the emergence of diverse recovery patterns.
When an error action $a_t$ is executed, the environment transitions to a failure state $s_{t+1}$, typically accompanied by execution feedback $o_\text{stderr}$ (e.g., \texttt{RuntimeError}).
In the subsequent steps, if the sampled intent reverts to $\mathcal{I}_{\text{correct}}$, we enforce a \textbf{recovery behavior}: the Step Generator is conditioned to analyze the preceding error feedback and synthesize a corrective action (e.g., debugging constraints or attempting an alternative command).
Crucially, since sampling is independent step-by-step, the system naturally allows for \textit{consecutive errors}.
This exposes the model to \textit{cascading failure modes}, teaching it not only to fix isolated mistakes but also to maintain diagnostic coherence across multi-turn, noisy interaction histories.
Note that these injected errors do not necessarily result in a failed trajectory, as the agent is required to recover from failures whenever a correct step is sampled in subsequent steps. 
These injected errors also do not reduce the success rate of completing task goals.

\subsection{Implementation and Data Statistics}
\label{subsec:statistics}

\textbf{Implementation Details.}
We utilized Claude-4.5-Sonnet~\citep{claudesonnet45} as the backbone model for agents in both Phase I and Phase II.
For trajectory collection, we implemented a minimal terminal-based agent framework, \agent, following the architecture of Terminus~\citep{terminus}.
The agent interacts with a Docker container via a raw bash shell, generating a ReAct-style~\citep{yao2022react} response at each turn (a reasoning trace followed by a bash command).
We deliberately exclude complex agent scaffolding (e.g., RAG or multi-agent debate) to ensure the collected data reflects the intrinsic reasoning capabilities required for fine-tuning.
A detailed task example and a visualization of the error-correction loop are in Appendix~\ref{append:example}.

\textbf{Dataset Statistics.}
Our synthesis pipeline yielded a final set of over $3,500$ verified environments, balanced across the From these environments, we curated a corpus of $3,291$ trajectories.
This dataset includes both successful solutions and unresolved attempts. 
As we will show in \cref{subsec:ablation}, these unresolved trajectories remain valuable for training.
Notably, these trajectories exhibit a \textit{long-horizon characteristic}, spanning an average of $25.5$ turns and $8,722$ tokens, reflecting the complexity of real-world system tasks.

\textbf{Tool Diversity.}
The dataset demonstrates broad coverage of terminal operations, utilizing a total of $420$ unique command-line tools.
These tools span $16$ functional domains, ranging from standard utilities (e.g., file and text processing) to specialized software in security forensics, network operations, and HPC environments.
\cref{fig:tool_distribution} illustrates the distribution of tool usage, highlighting the dataset's coverage of both foundational and domain-specific skills.

\begin{figure}[t!]
    \centering
    \includegraphics[width=0.45\textwidth]{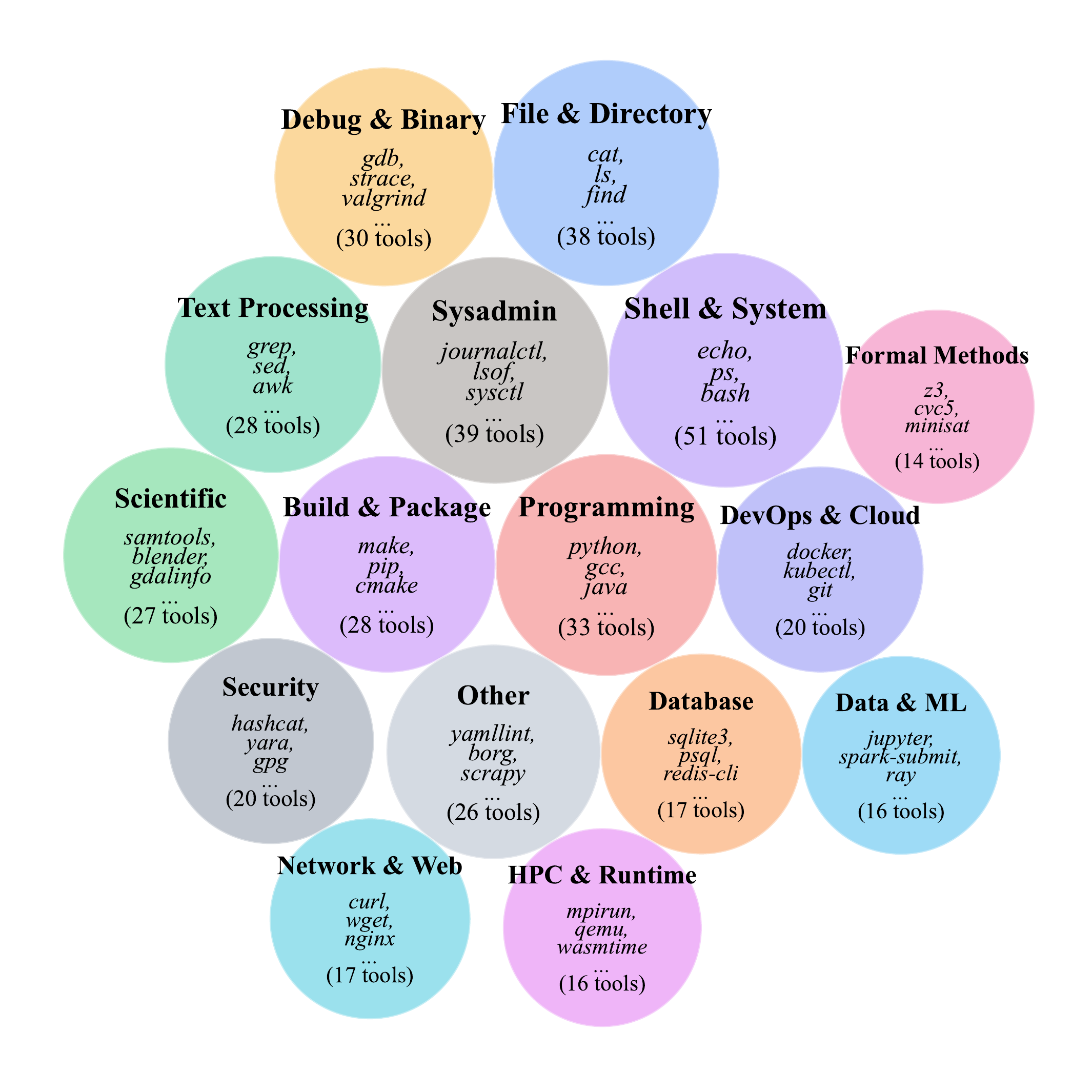}
    \caption{Distribution of $420$ command-line tools across $16$ functional categories. Bubble sizes are proportional to the logarithm of the number of tools in each category.}
    \label{fig:tool_distribution}
    \vspace{-.2in}
\end{figure}



\section{Experiments}
\label{sec:experiments}

We first validate the effectiveness of \sys by comparing our fine-tuned models against three categories of baselines: 1) proprietary frontier models, 2) general-purpose open-weights models, and 3) specialized models fine-tuned for terminal tasks.
Then, we answer the following research questions with rigorous ablation studies:
\underline{RQ1: Verifiable Environments vs. Simulation.} Does training on trajectories from physically verifiable environments (Phase I) yield better generalization than training on synthetic trajectories from simulated environments?
\underline{RQ2: Error-Correction vs. Standard Trajectory.} Does our controlled error injection strategy (Phase II) effectively improve the agent's self-correction capabilities compared to standard training on golden trajectories?
\underline{RQ3: The Value of Negative Trajectories.} How does the inclusion of \textit{unresolved error trajectories} (attempts that failed to reach a solution) during SFT affect the model's ability?


\subsection{Experiment Setup}
\label{subsec:setup}

\textbf{Models.}
For the student models, we select the \textit{Qwen-2.5-Coder-$32$B-Instruct}~\citep{hui2024qwen2} and \textit{Qwen3-32B}~\citep{yang2025qwen3}.
We choose these open-weight models as they represent the current state-of-the-art in code reasoning and general instruction following.
Training involves full-parameter supervised fine-tuning with a sequence length of $20,000$ for $5$ epochs to accommodate long interaction histories.
Hyperparameter details are provided in~\Cref{app:implementation}.

\textbf{Baselines.}
We evaluate TerminalBench against a comprehensive suite of models and frameworks, categorized into three distinct groups. 
First, we include proprietary frontiers and top-performing agent systems, Claude-4.5-sonnet~\cite{claudesonnet45},
Claude-3.7-Sonnet~\cite{claude37sonnet}, 
GPT-5~\cite{openai2025gpt5}, 
o4-mini~\cite{openai2025o3o4mini},
Claude Code~\cite{claude37sonnet}, Codex~\cite{openai2025codexproduct}, and Apex2, which establish the performance ceiling for autonomous terminal operations. 
Second, we assess agents built on general-purpose base models without task-specific fine-tuning, specifically Qwen3-235B-A30B~\citep{hui2024qwen2}, Qwen3-32B~\citep{hui2024qwen2}, Qwen-2.5-Coder-32B~\citep{yang2025qwen3}, and GPT-OSS-120B~\citep{openai2025gptoss}, paired with \agent~(described in \cref{subsec:statistics}).
Finally, we compare against specialized models that have been explicitly fine-tuned for terminal-based tasks using supervised fine-tuning or reinforcement learning, including Reptile~\citep{reptile2025}, LiteCoder-Terminal~\citep{litecoder}, and TerminalAgent~\citep{austin2025terminalbenchrl}.

\textbf{Benchmarks and Evaluation Metrics.}
To assess the agent's real-world capability, we employ TerminalBench~1.0~\citep{tbench2025} as our primary benchmark.
TerminalBench evaluates agents on a curated set of realistic terminal tasks proposed by various domain experts within isolated Docker containers.
Consistent with its standard evaluation protocol, we define the pass rate as our core metric.
For each task, the benchmark executes a dedicated verification script (\texttt{test.sh}), an agent is considered successful only if it passes all test cases defined in the verification script.
We set the generation temperature as $0.6$ for our fine-tuned models and base models.
To account for generation variance, we report the \textit{Average Pass@1} across three independent runs to ensure reliability.

\subsection{Main Comparisons}
\label{subsec:main}

\cref{tab:main_results} benchmarks our framework against both open-weights state-of-the-art (SOTA) and proprietary frontier models.
Our \sys-Qwen2.5-32B achieves a pass rate of $\textbf{31.3\%}$, establishing a new open-weights SOTA. 
It significantly outperforms existing fine-tuned agents such as Reptile ($18.9\%$) and TerminalAgent ($15.5\%$) by a substantial margin.
Crucially, compared to the untuned base models, \sys yields a nearly $25$\% performance gain.
Remarkably, our 32B model surpasses capable proprietary models like o4-mini ($20.0\%$) by $11.3$\%, despite having significantly fewer parameters and no access to other tools such as web browsing or memory.
While a gap remains compared to the apex of proprietary reasoning (e.g., Apex2 with Claude-4.5-Sonnet at $64.5\%$), \sys-Coder-32B reaches $73.1$\% of the performance of GPT-5-Codex. 
This result suggests that a specialized model, when trained on high-fidelity \textit{error-correction} data, can effectively approach the reasoning capabilities of much larger frontier models in domain-specific execution environments.

\begin{table}[t]
  \centering
  \caption{Comparison of \sys and baselines on Terminal-Bench 1.0. Results are sorted by pass rate in descending order. The * indicates results are reported by the authors.}
  \label{tab:main_results}
  \resizebox{\columnwidth}{!}{%
  \begin{tabular}{@{}l l c@{}}
    \toprule
    \textbf{Agent Framework} & \textbf{Base Model} & \textbf{Avg Pass Rate (\%)} \\
    \midrule
    \multicolumn{3}{l}{\textit{Proprietary Models}} \\
    \midrule
    Apex2          & Claude-4.5-Sonnet     & 64.5 $\pm$ 1.1 \\
    Codex CLI      & GPT-5-Codex           & 42.8 $\pm$ 4.2 \\
    Claude Code    & Claude-3.7-Sonnet     & 35.2 $\pm$ 2.6 \\
    Codex CLI      & o4-mini               & 20.0 $\pm$ 2.9 \\
    \midrule
    \multicolumn{3}{l}{\textit{General-purpose Base Models}} \\
    \midrule
    \agent         & GPT-OSS-120B          & 19.2 $\pm$ 2.4 \\
    \agent         & Qwen3-235B-A30B            & 15.4 $\pm$ 0.6 \\
    \agent         & Qwen3-32B             & 10.4 $\pm$ 1.2 \\
    \agent         & Qwen2.5-Coder-32B-Instruct     & 4.5 $\pm$ 0.7 \\
    Terminus 1     & Qwen3-235B-A30B            & 6.6 $\pm$ 2.8 \\
    \midrule
    \multicolumn{3}{l}{\textit{Fine-tuned Models}} \\
    \midrule
    \agent  & \textbf{\sys-Qwen2.5-Coder-32B} & \textbf{31.3 $\pm$ 1.8} \\
    \agent  & \textbf{\sys-Qwen3-32B}         & \textbf{27.5 $\pm$ 1.3} \\
    Reptile        & Devstral-2505-22B     & 18.9$^*$ \\
    Terminus 2     & LiteCoder-30a3b       & 18.8$^*$ \\
    Terminus 2     & TerminalAgent-32B     & 15.5 $\pm$ 2.2 \\
    \bottomrule
  \end{tabular}%
}
\end{table}

\subsection{Ablation Studies}
\label{subsec:ablation}

\textbf{RQ1: Verifiable Environments vs. Simulation.}
To quantify the necessity of physical execution, we investigate whether training on simulated observations yields comparable performance to training on real execution feedback.
We construct a baseline simulation dataset comprising $800$ trajectories.
To ensure a fair comparison, we maintain two identical experimental conditions to our main pipeline:
1) Tasks: The simulation uses the exact same task seeds and prompts as our verifiable dataset.
2) Strategy: We apply the same \textit{Error-Correction} strategy with a $20\%$ error injection probability.
The sole difference lies in the execution layer. 
Instead of executing commands in a Docker container, an \textit{Observation Synthesis Agent} (prompted with the current context) generates the hypothetical \texttt{stdout}, \texttt{stderr}.

As shown in~\cref{tab:sim_vs_real}, training on verifiable environments yields a performance advantage ($2.5$\%) over the simulation baseline.
With $800$ samples, the \texttt{Verifiable} model achieves a Pass Rate of $25.0\%$, improving the \texttt{Simulation} baseline by a relative improvement of $\approx 11\%$.
This result suggests that while simulation serves as a strong proxy, the fidelity of grounded execution provides the critical signals necessary to resolve complex edge cases.

\begin{table}

    \centering
    \small
    \caption{Comparison of average Pass@1 scores between synthesized observation (Simulation) and real observation (\sys), controlled for data size ($N=800$).}
    \label{tab:sim_vs_real}
  \resizebox{0.6\columnwidth}{!}{%
    \begin{tabular}{lc}
    \toprule
    \textbf{Observation Source} & \textbf{Avg Pass@1 (\%)} \\
    \midrule
    Simulation & 22.5 \\
    \textbf{\sys} & \textbf{25.0} \\
    \bottomrule
    \end{tabular}
    }
\vspace{-.1in}
\end{table}

To provide a rigorous explanation for the performance gap, we employ Claude-4.5-Sonnet to conduct an empirical audit of $50$ randomly sampled trajectories from the simulation dataset. 
Our analysis reveals that $26$\% of these trajectories contain observation errors with three systemic issues.
The most prevalent issue, \textit{Spurious Verbosity}, accounts for $53$\% of the identified errors. 
This occurs when the simulator hallucinates explicit confirmation outputs for commands that are natively silent, such as file redirections or heredocs. 
Additionally, $35$\% of errors involve \textit{Semantic Deviation}, where the simulator fails to faithfully replicate standard shell logic.
For instance, it may misattribute the root cause of a permission error or return incorrect exit codes. 
This deviation is particularly detrimental as it instills flawed diagnostic logic: the agent learns to associate specific error messages with incorrect root causes (e.g., applying \prompt{sudo} to fix a path error), resulting in ineffective debugging loops during real-world deployment.
Lastly, \textit{State Inconsistency} ($12$\%). For example, an agent first use \texttt{cat} to create a configuration file and receive a success signal, while in later steps, when the agent tries to read the config file, the simulator responds with \prompt{File not found}.
Such discontinuities break the logical chain of reasoning, making it impossible for the agent to debug effectively.

\textbf{RQ2: Error-Correction vs. Standard Trajectory.}
Next, we isolate the impact of our \textit{Error-Correction} strategy by comparing it against a model trained on standard trajectories.
We construct a standard baseline dataset with $800$ standard trajectories, which collected in the same verifiable environments but without active error injection.
Note that this baseline is not strictly error-free, the agent may still commit a few incidental mistakes.

As illustrated in~\cref{tab:standard_vs_error_injection}, our \sys model achieves a Pass@1 of $25.0$\%, consistently outperforming the baseline model ($21.8$\%) fine-tuned on standard trajectories. 
This performance gap indicates that high-density exposure to diverse failure modes generalizes effectively, improving end-to-end reasoning capabilities for small models.

To explain this gain, we conducted a granular analysis of the error distributions using the five-category taxonomy defined in \cref{subsec:trajectory_collection}.
We employ Claude-4.5-Sonnect to inspect $50$ trajectories from the baseline dataset and reveals a critical skewness in naturally occurring errors: they are dominated by \textit{Verification Failures} ($\approx 50\%$), where agents typically fail simply by omitting a final check, followed by \textit{Analysis Errors} ($\approx 19\%$).
Complex failure modes such as \textit{Hallucinations} and \textit{Requirement Violations} are rare.
In contrast, our strategy ($\epsilon=0.2$) \textit{actively forces} the agent to navigate a diverse spectrum of failure modes.
We further present a comparative case study in ~Appendix~\ref{append:qualitative_coq} to demonstrate the error-correction capabilities acquired through our training.

\begin{table}
    \centering
    \small
    \caption{Performance comparison between the baseline model (trained on standard trajectories) and our \sys method, controlled for data size ($N=800$).}
    \label{tab:standard_vs_error_injection}
  \resizebox{0.6\columnwidth}{!}{%
    \begin{tabular}{lc}
    \toprule
    \textbf{Training Strategy} & \textbf{Avg Pass@1 (\%)} \\
    \midrule
    Standard & 21.8 \\
    \textbf{\sys} & \textbf{25.0} \\
    \bottomrule
    \end{tabular}
}
\vspace{-.1in}

\end{table}

\textbf{RQ3: The Value of Negative Trajectories.}
Standard SFT protocols typically enforce a strict quality filter, retaining only trajectories that achieve a $100$\% pass rate on verification unit tests (e.g., reject sampling). 
We conduct an ablation study by varying the inclusion threshold $\tau$ based on the trajectory's unit test completion rate.
Specifically, we train Qwen-2.5-Coder-32B on trajectories where $\text{Test Pass Rate} \ge \tau$, with $\tau \in \{100\%, 50\%, 0\%\}$).
Counter-intuitively, as shown in \cref{tab:add_wrong_traj}, we observe that strict data filtering is detrimental in our setting: the model trained with $\tau=0\%$ achieves the highest performance, surpassing the strict $\tau=100\%$ baseline by a noticeable margin.

We attribute this performance gain to two factors. 
First, \textit{Task Complexity Coverage.} We hypothesize that trajectories achieving a $100$\% pass rate are statistically skewed towards easier tasks. 
Strictly filtering out imperfect runs inadvertently discards valuable exposure to complex, high-difficulty scenarios where the agent demonstrated valid reasoning logic but fell short of a perfect score.
Second, \textit{Local Recovery Supervision.} 
Due to our \textit{Error-Correction} mechanism, even a trajectory that ultimately fails the final evaluation, it still contains valid \textit{local} repairs of injected errors. 
These segments provide high-quality supervision for the mechanics of diagnosis and correction, independent of the global task outcome. 
Together, these factors suggest that imperfect data is essential for learning resilience in complex domains.

\begin{table}
    \centering
    \small
    \caption{Performance comparison when varying the inclusion threshold ($\tau$) based on trajectory unit test pass rates.}
    \label{tab:add_wrong_traj}
    \resizebox{0.9\columnwidth}{!}{%
    \begin{tabular}{lcc}
    \toprule
    \textbf{Inclusion Threshold ($\tau$)} & \textbf{Data Size} & \textbf{Avg Pass@1 (\%)} \\
    \midrule
    $\ge 100\%$ & $2,040$ & $25.0$ \\
    $\ge 50\%$ & $2,696$ & $26.3$ \\
    $\mathbf{\ge 0\%}$ & $\textbf{3,291}$ & $\textbf{31.3}$ \\
    \bottomrule
    \end{tabular}
    }
    \vspace{-.1in}
\end{table}

\section{Conclusion and Limitations}
\label{sec:conclusion}

In this work, we present a framework for synthesizing verifiable terminal environments and an error-injection distillation strategy. 
Our empirical results demonstrate that this approach closes the gap between open-weights and proprietary models. Despite these advancements, we acknowledge three key limitations that outline future directions.

First, our current training relies exclusively on SFT.
Since our environment provides deterministic verification signals via automated tests, a natural next step is to apply reinforcement learning. 
This would allow agents to explore novel solutions and learn from trial-and-error beyond the fixed distribution of the training data.
Second, we currently implement a simple agent without a memory component to mainly test the effectiveness of generated environments and trajectories. 
Future work could design more sophisticated agents with memory to leverage interaction histories.
Third, while our taxonomy spans diverse domains, our environments are inherently synthetic and isolated. 
They cannot fully replicate the stochasticity and scale, especially for real-world production systems (e.g., distributed clusters, high-concurrency traffic). 
Future research should investigate how agents trained on these controlled sandbox environments transfer to open-ended, large-scale infrastructure tasks.

\section*{Impact Statement}

This paper presents \sys, a framework that improves the reliability of autonomous agents in software engineering and system administration. 
By enabling open-source models to effectively use terminal tools, we make advanced DevOps capabilities more accessible to the community, lowering the barrier for complex software tasks.

However, we acknowledge that giving agents the power to execute terminal commands comes with risks. 
\textbf{Safety Risks:} An autonomous agent might accidentally cause data loss or system crashes (e.g., deleting the wrong files). Our focus error correction helps reduce this risk. By teaching agents to identify and fix their own mistakes, we prevent simple errors from leading to larger failures.
\textbf{Malicious Use:} The techniques described could theoretically be misused to automate cyber-attacks. 
To address these risks, we emphasize that such agents must be deployed in isolated environments (e.g., Docker containers) and always under human supervision.

\bibliography{ref}
\bibliographystyle{icml2026}

\newpage

\appendix
\onecolumn
\crefalias{section}{appendix}
\crefname{appendix}{Appendix}{Appendices}
\Crefname{appendix}{Appendix}{Appendices}

\section{Detailed Task Categories}
\begin{table*}[t!]
    \centering
    \small
    \caption{\textbf{Hierarchical Domain Taxonomy.} Our dataset covers three tiers spanning 11 sub-categories, ranging from low-level infrastructure to high-level specialized development. We list representative competencies and specific task seeds for each category.}
    \label{tab:taxonomy}
    \renewcommand{\arraystretch}{1.3} 
    \begin{tabularx}{\textwidth}{@{} p{0.20\textwidth} X p{0.35\textwidth} @{}}
        \toprule
        \textbf{Sub-Category} & \textbf{Competencies (Goals \& Skills)} & \textbf{Representative Task Seeds} \\
        \midrule
        \multicolumn{3}{l}{\textit{\textbf{Tier I: Infrastructure \& Core Systems}}} \\
        \cmidrule(l){1-3}
        1.1 Software Build \& \newline Compilation & 
        \textbf{Goal:} Resolve dependency hell, linker errors, and cross-compilation issues. \newline 
        \textbf{Skills:} \texttt{gcc}, \texttt{cmake}, \texttt{makefile}, \texttt{rustc}, \texttt{autotools}, \texttt{ld}. & 
        \texttt{gcc\_cannot\_find\_library}, \newline \texttt{cmake\_cuda\_toolkit\_not\_found}, \newline \texttt{rust\_cargo\_linker\_failure}, \newline \texttt{makefile\_missing\_separator} \\
        
        1.2 System Administration \& DevOps & 
        \textbf{Goal:} Manage containers, orchestrate clusters, and debug system services. \newline 
        \textbf{Skills:} \texttt{docker}, \texttt{kubernetes}, \texttt{systemd}, \texttt{nginx}, \texttt{terraform}. & 
        \texttt{kubernetes\_pod\_crashloop\_backoff}, \newline \texttt{docker\_layer\_caching\_broken}, \newline \texttt{nginx\_upstream\_keepalive\_connections}, \newline \texttt{systemd\_unit\_fails\_on\_boot} \\
        
        1.3 Security, Forensics \& \newline Reverse Engineering & 
        \textbf{Goal:} Exploit analysis, binary reversing, and digital forensics. \newline 
        \textbf{Skills:} \texttt{ghidra}, \texttt{wireshark}, \texttt{metasploit}, \texttt{gdb}, \texttt{volatility}. & 
        \texttt{buffer\_overflow\_exploit\_development}, \newline \texttt{wireshark\_pcap\_analysis}, \newline \texttt{ghidra\_reverse\_engineering\_workflow}, \newline \texttt{sqlmap\_automated\_injection} \\
        
        \midrule
        \multicolumn{3}{l}{\textit{\textbf{Tier II: Data \& Algorithm Applications}}} \\
        \cmidrule(l){1-3}
        2.1 Data Processing \& \newline ETL & 
        \textbf{Goal:} Transform large-scale datasets and handle schema evolution. \newline 
        \textbf{Skills:} \texttt{pandas}, \texttt{spark}, \texttt{kafka}, \texttt{sql}, \texttt{parquet}, \texttt{jq}. & 
        \texttt{parquet\_schema\_evolution}, \newline \texttt{kafka\_avro\_schema\_registry}, \newline \texttt{pandas\_merge\_asof\_timeseries}, \newline \texttt{spark\_dataframe\_schema\_inference} \\
        
        2.2 Machine Learning \& \newline MLOps & 
        \textbf{Goal:} Debug training instability, optimize inference, and manage pipelines. \newline 
        \textbf{Skills:} \texttt{pytorch}, \texttt{cuda}, \texttt{huggingface}, \texttt{scikit-learn}. & 
        \texttt{cuda\_out\_of\_memory}, \newline \texttt{pytorch\_dataloader\_workers}, \newline \texttt{transformers\_token\_limit}, \newline \texttt{gradient\_explosion\_detection} \\
        
        2.3 Algorithms \& \newline Logic Puzzles & 
        \textbf{Goal:} Implement classic algorithms and solve competitive programming tasks. \newline 
        \textbf{Skills:} Graph theory, Dynamic Programming, Search (A*), CSP. & 
        \texttt{n\_queens\_placement}, \newline \texttt{shortest\_path\_dijkstra}, \newline \texttt{alpha\_beta\_pruning\_minimax}, \newline \texttt{knapsack\_dynamic\_programming} \\
        
        \midrule
        \multicolumn{3}{l}{\textit{\textbf{Tier III: Specialized Domains \& Advanced Development}}} \\
        \cmidrule(l){1-3}
        3.1 Software Dev, \newline Porting \& Bug Fixing & 
        \textbf{Goal:} Full-stack debugging, framework configuration, and legacy porting. \newline 
        \textbf{Skills:} \texttt{react}, \texttt{django}, \texttt{git}, \texttt{rest-api}, \texttt{ci/cd}. & 
        \texttt{react\_hooks\_dependency\_array}, \newline \texttt{django\_csrf\_token\_missing}, \newline \texttt{git\_merge\_conflict\_binary}, \newline \texttt{rest\_api\_pagination\_cursor} \\
        
        3.2 Scientific \& \newline Domain Computing & 
        \textbf{Goal:} Computational biology, chemistry simulations, and statistical modeling. \newline 
        \textbf{Skills:} \texttt{bioconductor}, \texttt{rdkit}, \texttt{gromacs}, \texttt{stan}, \texttt{numpy}. & 
        \texttt{rdkit\_mol\_sanitization\_error}, \newline \texttt{stan\_divergent\_transitions}, \newline \texttt{bioconductor\_annotation\_mismatch}, \newline \texttt{gromacs\_topology\_atom\_mismatch} \\
        
        3.3 Interactive \newline Environments & 
        \textbf{Goal:} Handle real-time protocols, REPLs, and interactive debugging sessions. \newline 
        \textbf{Skills:} \texttt{websocket}, \texttt{ssh}, \texttt{gdb-interactive}, \texttt{jupyter}. & 
        \texttt{websocket\_chat\_server}, \newline \texttt{interactive\_sql\_repl}, \newline \texttt{gdb\_interactive\_debugging}, \newline \texttt{jupyter\_kernel\_connection} \\
        
        3.4 Distributed \& \newline Parallel Computing & 
        \textbf{Goal:} Debug race conditions, deadlocks, and distributed consensus issues. \newline 
        \textbf{Skills:} \texttt{mpi}, \texttt{openmp}, \texttt{ray}, \texttt{dask}, \texttt{slurm}. & 
        \texttt{mpi\_deadlock\_collective\_ops}, \newline \texttt{spark\_executor\_out\_of\_memory}, \newline \texttt{ray\_actor\_died\_unexpectedly}, \newline \texttt{openmp\_race\_condition\_shared} \\
        
        3.5 Formal Verification \newline \& Graphics & 
        \textbf{Goal:} Theorem proving, SAT solving, and rendering pipeline debugging. \newline 
        \textbf{Skills:} \texttt{coq}, \texttt{z3}, \texttt{opengl}, \texttt{vulkan}, \texttt{blender}. & 
        \texttt{coq\_universe\_inconsistency}, \newline \texttt{opengl\_context\_creation\_headless}, \newline \texttt{z3\_timeout\_optimization}, \newline \texttt{vulkan\_validation\_layer\_error} \\
        \bottomrule
    \end{tabularx}
\end{table*}

~\Cref{tab:taxonomy} shows the task taxonomy. 

\begin{figure}[t!]
    \centering
    \includegraphics[width=0.7\textwidth]{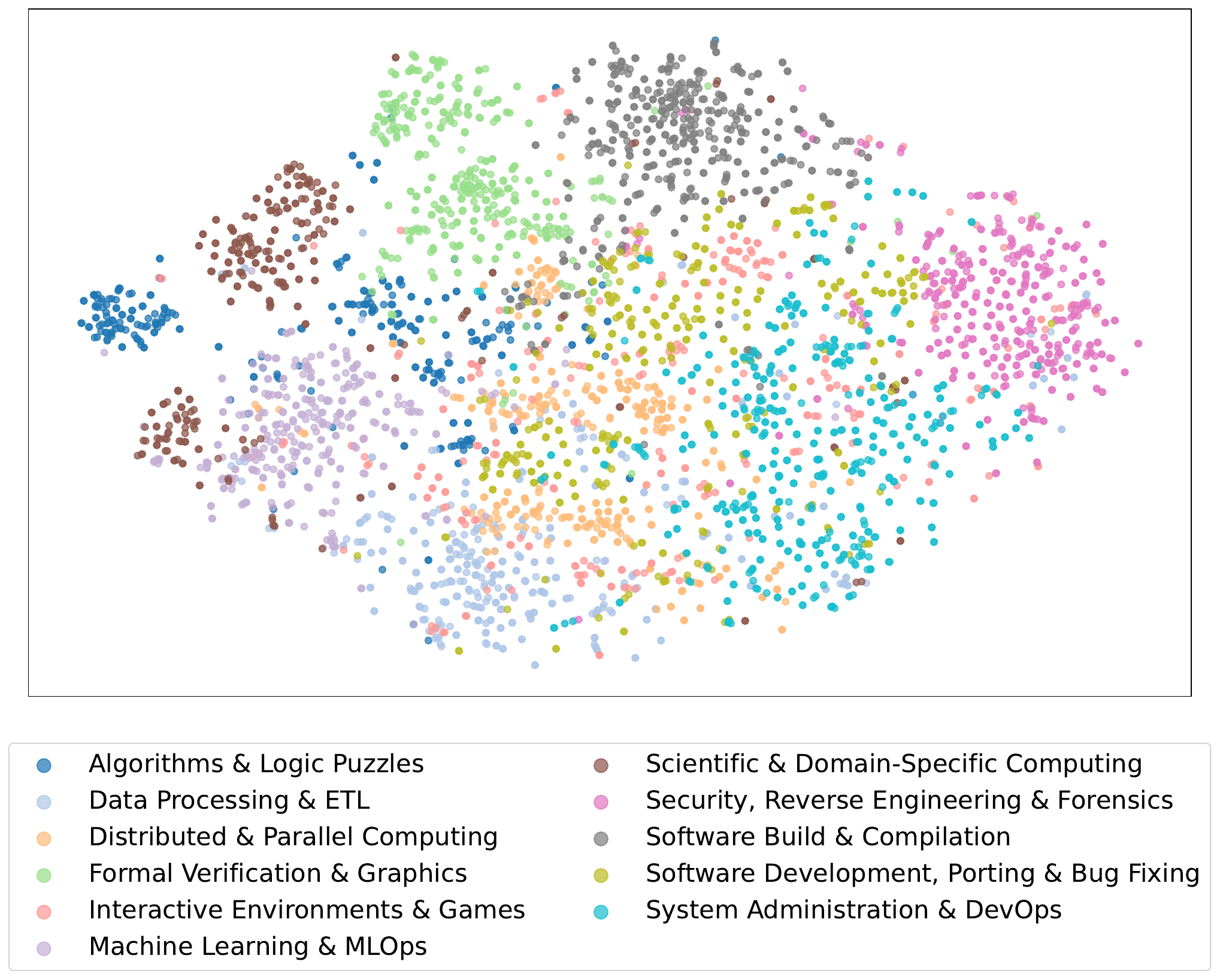}
    \caption{t-SNE visualization of $\approx 3,500$ tasks across $11$ categories, showing semantic clustering by task type.}
    \label{fig:task_diversity}
\end{figure}

\section{Implementation Details}
\label{app:implementation}
Our training pipeline is built upon the \texttt{LLaMA-Factory} framework.
We perform full-parameter supervised fine-tuning (SFT) on the base model using the synthetically generated dataset.
To accommodate the extensive context requirements of multi-turn terminal interactions, we set the maximum sequence length to $20,000$ tokens. 
The training is optimized using AdamW with a learning rate of $5.0 \times 10^{-6}$, employing a cosine learning rate scheduler with a warmup ratio of $0.1$ and zero weight decay. 
We train the model for $5$ epochs with a per-device batch size of $1$ and $8$ gradient accumulation steps. 
All experiments were conducted on the compute nodes equipped with $8\times$ AMD MI325X GPUs.
To ensure memory efficiency and scalability across these accelerators, we utilize DeepSpeed ZeRO-3 with BF16 precision enabled.

\section{Task and Error-Correction Trajectory}
\label{append:example}

\subsection{Synthetic Dataset Samples}
\label{append:task-example}
In this case study, we present a representative synthetic task directory,
to illustrate the structure and content of our benchmark.

\paragraph{Task Description.}
This task targets macOS automation using AppleScript.
The high-level goal is to locate all application bundles (.app packages) within the
\texttt{/Applications} directory (including subdirectories) and extract their version information from Info.plist files. 
The task requires handling edge cases such as: (1) Application bundles nested in subdirectories; (2) Missing or malformed Info.plist files; (3) Applications without readable version information.

The solution must generate a JSON file at \texttt{/tmp/app\_inventory.json} containing: 
(1) \texttt{total\_apps}: Total number of .app bundles found;
(2) \texttt{apps\_with\_version}: Number of apps with readable version information
(3) \texttt{apps\_without\_version}: Number of apps where version info couldn't be extracted.

\paragraph{Directory Structure.}
The directory is organized as follows:

\begin{lstlisting}[basicstyle=\ttfamily\small, breaklines=true, frame=single, columns=flexible, keepspaces=true]
|-- task.yaml              # Task description and metadata
|-- Dockerfile             # Docker environment configuration
|-- docker-compose.yaml    # Docker Compose configuration
|-- run-tests.sh           # Test execution script
|-- Applications/          # Test data directory
|   |-- TextEditor.app/
|   |   |-- Contents/
|   |   |   |-- Info.plist
|   |-- Utils/
|   |   |-- Calculator.app/
|   |   |   |-- Contents/
|   |   |   |   |-- Info.plist
|   |-- DevTools/
|   |   |-- CodeEditor.app/
|   |   |   |-- Contents/
|   |   |   |   |-- Info.plist
|   |   |-- Terminal.app/
|   |   |   |-- Contents/
|   |   |   |   |-- Info.plist
|-- tests/
    |-- test_outputs.py    # Automated test suite
\end{lstlisting}

\paragraph{Evaluation Framework.}
The evaluation is performed through a comprehensive pytest-based test suite
(\texttt{tests/test\_outputs.py}) that validates: (1) Output file existence at the specified path; (2) Valid JSON format; (3) Presence of all three required fields; (4) Exactly three fields (no extra fields); (4) Correct data types (all integers);
(5) Accurate count of total applications (must equal 4).

\subsection{An Error-Correction Example in Training Trajectories}
We present a representative example illustrating the \textit{error analysis and correction loop} in our generated trajectories.

\begin{framed}
\prompt{
\user \ (Goal):

Solve the SMT constraints in /workspace/password\_policies.smt2 using CVC5. Goal: Determine satisfiability and extract a valid password model if one exists. Output: Save the result to /workspace/result.txt in the exact format: satisfiable: yes followed by password: <value> on the next line (or just satisfiable: no if impossible).

\robot \ (Action 1):

...

\tool \ (Output 1):

...

\robot \ (Action 2):

...

\tool \ (Output 2):

...

\robot \ (Action 3):

Looking at the current state, I have:
\begin{itemize}[leftmargin=*, noitemsep, topsep=0pt]
    \item Confirmed the \texttt{/workspace/password\_policies.smt2} file exists and contains valid SMT-LIB2 constraints
    \item Verified that CVC5 is available at \texttt{/usr/local/bin/cvc5}
    \item Understood the password constraints from the SMT file
\end{itemize}

The next logical step is to run CVC5 on the SMT file to determine if the constraints are satisfiable and, if so, get a model (example password) that satisfies all constraints.

\vspace{0.5em}
<tool\_call>\{cvc5 --lang smt2 --strings-exp /workspace/password\_policies.smt2\}\\</tool\_call>

{\color{blue} $\Rightarrow$ This command is intentionally incorrect to inject an error.}

\vspace{0.5em}

\tool \ (Output 3):

{\color{red} Parse Error: /workspace/password\_policies.smt2:6.33: Symbol '>=' not declared as a variable}

\robot \ (Action 4):

The error persists - CVC5 is still reporting that '>=' is not declared as a variable. This is happening because the SMT-LIB2 file declares the logic as `QF\_S` (quantifier-free strings only), but then uses integer comparison operators (`>=`, `<=`) with `str.len`, which requires integer arithmetic support.

The issue is that QF\_S logic doesn't include integer arithmetic. The correct logic should be `QF\_SLIA` (Quantifier-Free Strings and Linear Integer Arithmetic) or just let CVC5 auto-detect it.

{\color{blue} $\Rightarrow$ In this step, it reflects and refines previous error command.}

However, I notice the task says \"The file contains a complete, valid SMT-LIB2 problem statement\" - this suggests the file should work as-is. 
Let me check if there's a way to make CVC5 more lenient or if I need to use a different approach.

Let me try running CVC5 without specifying the logic strictly, by using the `--force-logic` option to override the declared logic:

<tool\_call>\{cvc5 --force-logic=QF\_SLIA /workspace/password\_policies.smt2\}\\</tool\_call>

}

\end{framed}

\section{Case Study: Learned Error-Correction Capabilities in \sys}
\label{append:qualitative_coq}

To demonstrate the robust error-correction capabilities of \sys, we conduct a comparative case study against the baseline model (Qwen2.5-Coder-32B) and our fine-tuned model (\sys-Qwen2.5-Coder-32B) on a challenging formal verification task: proving the commutativity of addition (\texttt{n + m = m + n}) in Coq.

\textbf{Baseline Failure: Premature Conclusion.}
As summarized in \cref{table:error-correction-compare}, the baseline model attempts a standard proof using induction and reflexivity.
However, it suffers from a silent failure mode: upon generating the code, it assumes correctness without verifying the generation of the compiled object file (`.vo`).
Lacking a verification loop, it fails to detect that the proof is mathematically incomplete, resulting in a failed trajectory despite a confident termination.

\textbf{\sys Success: Reasoning-Driven Recovery.}
In contrast, \sys exhibits a distinct \textit{iterative debugging} behavior.
In the initial attempt, \sys encounters a compilation error: \texttt{Unable to unify "m + 0" with "0 + m"}.
Instead of hallucinating or giving up, the model analyzes the stderr output and correctly identifies the root cause: \texttt{reflexivity} only handles definitional equality, whereas the goal requires propositional equality via rewriting.
Guided by this diagnosis, \sys applies the necessary lemmas (\texttt{rewrite plus\_0\_r} and \texttt{plus\_Snm}) across two correction steps, ultimately producing a verified, compiling proof.

This comparison highlights that our error-injection training has endowed \sys with the critical ability to treat execution feedback as a navigational signal rather than a terminal state.

\begin{table}
\label{table:error-correction-compare}
    \centering
    \small
    \caption{\textbf{Qualitative Comparison on Coq Theorem Proving.} The baseline model fails due to a lack of verification, \sys successfully navigates multiple error states through iterative diagnosis and correction.}
    \label{tab:case_study_coq}
    \renewcommand{\arraystretch}{1.2}
    \begin{tabular}{l p{0.38\textwidth} p{0.42\textwidth}}
        \toprule
        \textbf{Aspect} & \textbf{Baseline (Qwen2.5-Coder-32B)} & \textbf{\sys (Ours)} \\
        \midrule
        \textbf{Behavior} & \textit{One-shot attempt.} Wrote a proof and exited immediately without checks. & \textit{Iterative refinement.} Engaged in a multi-turn debug loop (20 episodes). \\
        \midrule
        \textbf{Error Handling} & \textbf{None.} Ignored potential compilation issues and assumed success. & \textbf{Active Diagnosis.} Analyzed stderr (\texttt{Unable to unify...}) to distinguish definitional vs. propositional equality. \\
        \midrule
        \textbf{Key Action} & Failed to verify the output artifact (`.vo` file). & Applied specific fixes: \texttt{rewrite plus\_0\_r} and \texttt{rewrite plus\_Snm}. \\
        \midrule
        \textbf{Outcome} & \textcolor{red}{\textbf{Failed}} (Incomplete proof) & \textcolor{green}{\textbf{Success}} (Verified proof) \\
        \bottomrule
    \end{tabular}
\end{table}


\end{document}